\documentclass[conference]{IEEEtran}
\IEEEoverridecommandlockouts
\usepackage{amsmath,amssymb,amsfonts}
\usepackage{algorithmic}
\usepackage{graphicx}
\usepackage{textcomp}
\usepackage{amsfonts}
\usepackage[ruled,vlined]{algorithm2e}
\usepackage{flushend}

\usepackage[backend=bibtex]{biblatex}
\bibliography{refbib_v2}

\usepackage[usernames,dvipsnames]{xcolor}

\def\BibTeX{{\rm B\kern-.05em{\sc i\kern-.025em b}\kern-.08em
    T\kern-.1667em\lower.7ex\hbox{E}\kern-.125emX}}
\begin{document}

\title{Automated stability testing of elastic rods with helical centerlines using a robotic system\
\thanks{We thank the funding and support from the National Science Foundation under Grant Numbers IIS-1925360, CAREER-2047663, and CMMI-2101751.}
}

\author{Dezhong Tong$^{1}$, Andy Borum$^{2}$, Mohammad Khalid Jawed$^{1}$
\thanks{$^{1}$Department of Mechanical \& Aerospace Engineering, University of California, Los Angeles, 420 Westwood Plaza, Los Angeles, CA 90095}
\thanks{$^{2}$Department of Engineering, Hofstra University, Hempstead, NY 11549}
}

\maketitle

\begin{abstract}
Experimental analysis of the mechanics of a deformable object, and particularly its stability, requires repetitive testing and, depending on the complexity of the object's shape, a testing setup that can manipulate many degrees of freedom at the object's boundary.  Motivated by recent advancements in robotic manipulation of deformable objects, this paper addresses these challenges by constructing a method for automated stability testing of a slender elastic rod --- a canonical example of a deformable object --- using a robotic system.  We focus on rod configurations with helical centerlines since the stability of a helical rod can be described using only three parameters, but experimentally determining the stability requires manipulation of both the position and orientation at one end of the rod, which is not possible using traditional experimental methods that only actuate a limited number of degrees of freedom.  Using a recent geometric characterization of stability for helical rods, we construct and implement a manipulation scheme to explore the space of stable helices, and we use a vision system to detect the onset of instabilities within this space. The experimental results obtained by our automated testing system show good agreement with numerical simulations of elastic rods in helical configurations. The methods described in this paper lay the groundwork for automation to grow within the field of experimental mechanics.

\end{abstract}


\section{Introduction}
A deformable elastic rod can conform to an infinite number of equilibrium configurations, even when its boundary conditions are fixed.  Determining which of these configurations are stable and which are unstable is a quintessential problem in mechanics. However, experimental analysis of these configurations requires repetitive testing and an experimental setup that can actuate up to six degrees of freedom at the rod's ends. In prior work, traditional experiment platform have only controlled a small number (e.g., one or two) of degrees of freedom at the ends of a rod~\cite{lazarus2013contorting,thompson1999helix}. In Ref.~\cite{lazarus2013contorting}, an experimental platform with two DOFs is used to observe the rod's behavior under contorting. Thompson et. al. implemented a platform controlling two DOFs to study post-buckling of a flexible rod with torsion~\cite{thompson1999helix}. The number of controllable DOFs often restricts the configurations of the rod that can be studied in experiments. In this paper, we address these challenges by constructing an automated testing method using a robotic system to determine when an elastic rod becomes unstable. Although the methods described in this paper could be applied to a rod in any configuration, we focus on rod configurations whose centerline is a helix.  This choice is motivated by a recent characterization of the set of all helical rod configurations that are stable~\cite{borum2020helix}, and we exploit this characterization in our automated testing method.  Experimental analysis of this stability problem has not been completed in previous work due to the repetitive nature of the testing procedure and the complexity of controlling both the position and orientation at the rod's end.  These issues are overcome by employing a robot to perform the stability experiments.

\begin{figure*}[h!]
\centerline{\includegraphics[width =2\columnwidth]{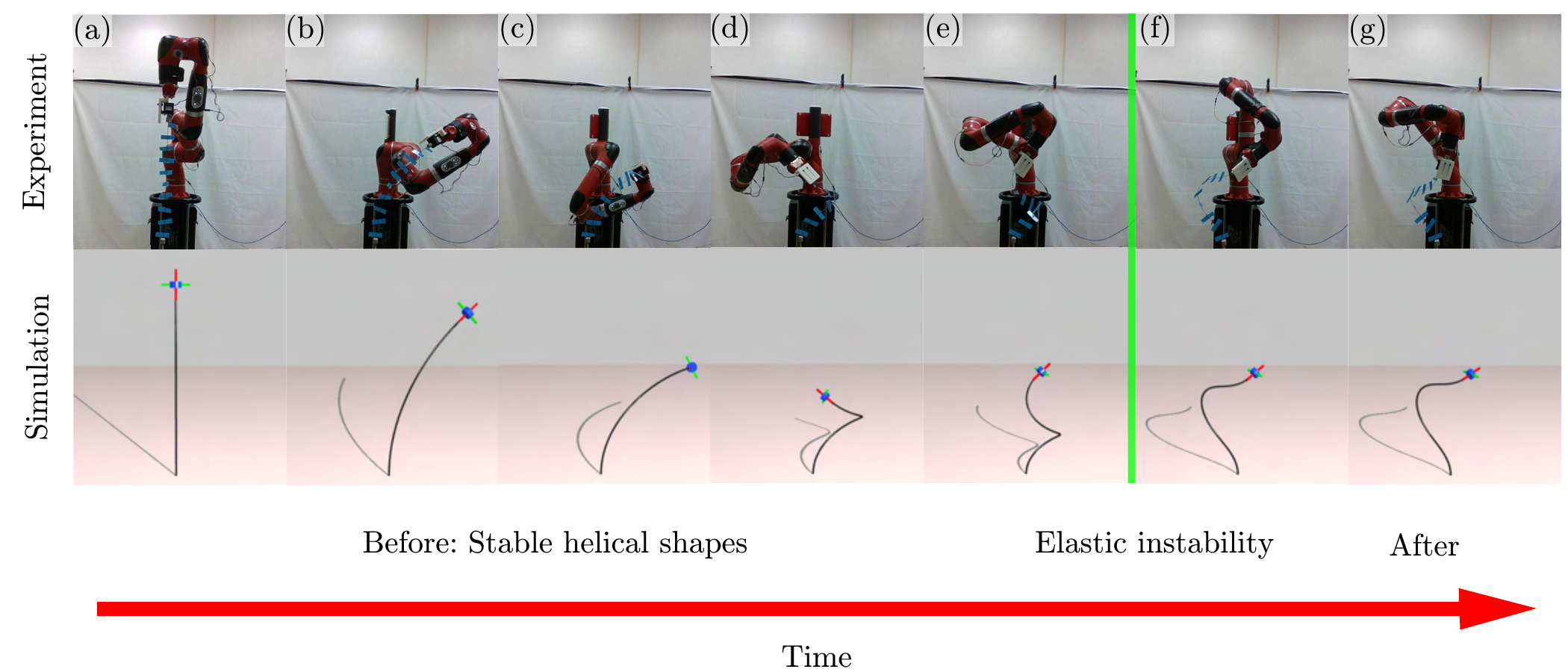}}
\caption{Snapshots of an elastic rod manipulated by a robot from stable helical configurations to non-helical configurations after an instability. (a-e) Stable helical configurations; (f-g) non-helical configurations after an instability; elastic instability happens at the vertical green line.
}
\label{fig::fig1}
\end{figure*}

Fig.~\ref{fig::fig1} provides one example of an experiment conducted by our automated testing system.  The  elastic rod, denoted with blue markers along its length, deforms with a helical centerline as the collaborative robot manipulates one end of the rod in Fig.~\ref{fig::fig1}(a-e).  Between Fig.~\ref{fig::fig1}(e) and (f), an instability occurs, and the rod jumps to another configuration that does not have a helical centerline.  The experiment concludes in Fig.~\ref{fig::fig1}(g).  Simulated configurations of the rod with identical boundary conditions, based on the Discrete Elastic Rod formulation~\cite{bergou2010discrete, jawed2018primer}, are shown beneath the experimental images.

A robot manipulating a deformable object, such as in Fig.~\ref{fig::fig1}, faces many challenges that do not occur when manipulating rigid objects.  The problem of deformable object manipulation has therefore received considerable attention in the robotics literature~\cite{henrich2012robot}.  In this prior work, the goal is often to deform an object, such as the elastic rod considered in this paper, from a starting configuration into a specific goal configuration~\cite{lamiraux2001planning,moll2006path}.  Other constraints, such as limiting the object's deformation, avoiding unstable configurations, and avoiding self-collisions, are often included~\cite{tanner2006mobile,sintov2020motion,bretl2014quasi}. Problems involving the manipulation of elastic rods arise in a variety of applications in different engineering fields, for example, robot cutting with a hot wire~\cite{duenser2020robocut}, magnetically guided rods for medical applications~\cite{kratchman2016guiding}, and the development of concentric tube robots~\cite{gilbert2015elastic}. The methods in this paper are strongly motivated by this previous work on robotic manipulation.  However, rather than using a robot to deform an object into a goal configuration while avoiding undesirable phenomena (such as an instability), our goal is to incite an instability in order to gain information regarding the mechanics of the object being manipulated.

In addition to the robotic system, a key component of our automated stability testing method is a vision system that detects when the rod becomes unstable.  Like manipulation, visual tracking of deformable objects has previously been studied in the robotics literature~\cite{javdani2011modeling}, with applications in robotic surgery~\cite{jackson2017real}.  In contrast to our work, the goal of these previous studies was to estimate the rod's current configuration using, for example, an image of the rod.  In this paper, we also estimate the rod's configuration using an image provided by a camera.  However, we then measure the error between this detected configuration and the expected configuration.  As we will show, a sudden jump in this error can be used as an automated method to determine when an instability occurs.

The elastic rods considered in this paper are examples of flexible slender structures, which are commonly encountered in our daily lives.  The study of their equilibrium configurations has many practical applications, including polymers, bacterial fibers, DNA, and plant growth~\cite{wada2012hierarchical,goldstein2000bistable,jawed2015propulsion,vologodskii1979statistical}. Beyond equilibrium, determining which of these configurations are stable and which are unstable has been a topic of much interest in the mechanics community~\cite{moulton2018stable,kumar2010generalized}.  Configurations whose centerline is a helix have received particular attention, dating back to Kirchhoff's seminal work in 1859, when he showed that an initially straight, inextensible, unshearable, isotropic, and uniform rod can have a helical centerline under appropriate boundary conditions.  It was later shown that the stability of these helical configurations is determined by only three parameters: the centerline's curvature, the centerline's torsion, and the twisting moment applied to the rod~\cite{borum2020helix}.  Furthermore, within this three-dimensional parameter space, the set of helical configurations that are stable is star-convex.  The manipulation scheme that we use to incite instabilities in the elastic rod is motivated by this geometric property of the set of stable helical configurations.  Finally, to validate the results of our automated testing method, we compare the stability measurements collected by our robotic system with simulations of helical elastic rods, which are based on the Discrete Elastic Rod formulation~\cite{bergou2010discrete,jawed2018primer}. The mechanical information obtained from these robotic experiments could also be used to develop a refined mechanical model. In Ref.~\cite{logarzo2021smart}, machine learning is used to construct smart constitutive laws of materials, which could be deployed in traditional finite element analysis to study the deformation of complex structures. In future work, the robotic system described in this paper can be combined with machine learning to use information collected from mechanical experiments to obtain a better understanding of complex structures.

The primary contributions of our work are outlined below.
\begin{itemize}
	\item We describe an automated procedure for testing the stability of elastic rod configurations using a robotic system.
	\item We use this system to determine when rod configurations with helical centerlines become unstable.
	\item We use numerical simulations to validate both our experimental results and the feasibility of using a robot to perform mechanics experiments.
\end{itemize}

The paper is organized as follows. In Section~\ref{sec::ps}, we describe the parameter space of rods with helical centerlines, the numerical simulation framework, and the manipulation scheme to explore the set of stable helices.  Section~\ref{sec::robot} describes the robotic system that is used to perform the experiments, and Section~\ref{sec::exp} gives the result of the experiments. Section~\ref{sec::conclusion} provides concluding remarks and directions for future work.

\section{Exploration and simulation of helical rods}
\label{sec::ps}
This section describes a parameterized space for helical rod configurations and proposes a scheme to explore the subset of this space corresponding to configurations that are stable. We first describe the coordinates of this parameter space: the rod's curvature, torsion, and twisting moment. We then describe the details of our numerical simulation to predict the stability of helical rod configurations. Finally, we describe a manipulation scheme to incite instabilities in helical rods, which is later implemented in Section~\ref{sec::robot} on the robotic system.


\subsection{Parameter space of helical rod configurations}
In this section, we present the necessary background from Ref. \cite{borum2020helix}. This work showed that the set of all stable helical configurations of an initially straight, inextensible, unshearable, isotropic, and uniform elastic rod having length $L$ can be parameterized by the centerline's curvature $\kappa \ge 0$ and torsion $\tau$, and the twisting moment $\omega$ applied to the rod.  Knowledge of these three parameters is sufficient to determine if the corresponding helical rod is stable.  We can therefore visualize the stability of all helical configurations within a three-dimensional parameter space having axes $\kappa$, $\tau$, and $\omega$.

Each point in this parameter space corresponds to a helical rod configuration that is in equilibrium.  However, only certain points in this space will correspond to helical configurations that are stable, i.e., configurations that minimize elastic potential energy.  The elastic potential energy is comprised of two terms: the bending energy and the twisting energy.  We note that the energy associated with axial stretching is neglected, as is potential energy due to external forces such as gravity.  The relative weights of bending and twisting energy are determined by the stiffness ratio $c = k_t/k_b = 1/(1 + \nu)$, where $k_b$ is the bending stiffness, $k_t$ is the twisting stiffness, and $\nu$ is the Poisson's ratio.  The ratio $c$, however, does not affect stability within the $\kappa$-$\tau$-$\omega$ parameter space.

A geometric property of the stable subset, derived in Ref. \cite{borum2020helix}, states that the set of points in the $\kappa$-$\tau$-$\omega$ parameter space corresponding to stable configurations is star-convex.  This means that each ray extending from the origin in the $\kappa$-$\tau$-$\omega$ parameter space intersects the boundary separating stable and unstable helices exactly once.  In the following sections, we will use this property to construct a manipulation scheme to incite instabilities in our automated testing procedure.

\subsection{Numerical framework}
A numerical framework based on Discrete Elastic Rod (DER)~\cite{bergou2010discrete, jawed2018primer} is used to simulate a rod under various boundary conditions and generate a corresponding robotic trajectory for motion planning. DER is a simulation tool developed in the computer graphics community and has recently gained traction in engineering as a predictive tool. Since the physical accuracy of DER has been validated in different studies~\cite{jawed2014coiling,jawed2015propulsion,baek2018form}, we use this method to study formation of helices in elastic rods and their stability. A brief description of DER follows; a tutorial exposition can be found in Ref.~\cite{jawed2018primer}.

\begin{figure}[t!]
\centerline{\includegraphics[width =\columnwidth]{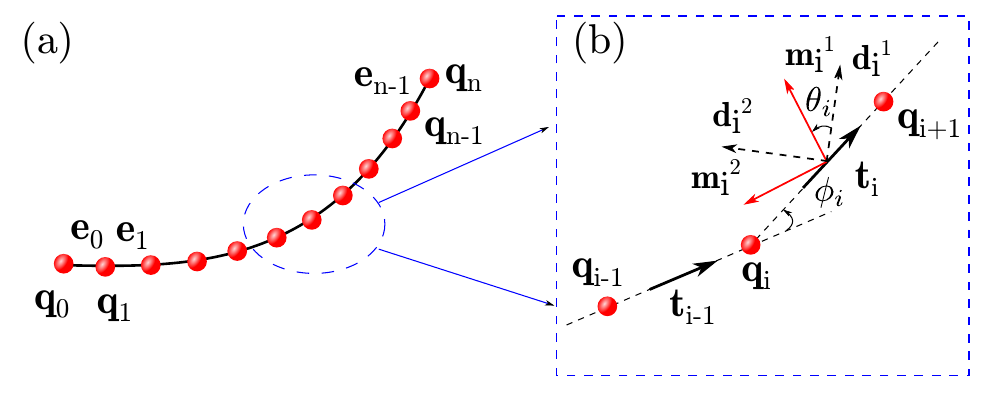}}
\caption{(a) Discrete model of an elastic rod; (b) Material frame, reference frame, and turning angle.}
\label{fig::DER}
\end{figure}

Fig.~\ref{fig::DER}(a) schematically represents the discrete representation of a rod in DER. The rod is discretized into $n+1$ nodes $\mathbf{q}_i$ $(0 \leq i \leq n)$, and $n$ edges $\mathbf{e}_i = \mathbf{q}_{i+1} - \mathbf{q}_{i}$ $(0 \leq i \leq n-1)$ as shown in Fig.~\ref{fig::DER}(a). Each node $\mathbf{q}_i$ has three degrees of freedom: positions along $x$, $y$ and $z$ axes; and each edge $\mathbf{e}_i$ has one degree of freedom -- twist angle $\theta_i$ -- that represents the orientation of the material frame $[\mathbf{m}_{i}^1, \mathbf{m}_{i}^2, \mathbf{t}_{i}]$ with respect to the reference frame $[\mathbf{d}_{i}^1, \mathbf{d}_{i}^2, \mathbf{t}_{i}]$. The reference frame $[\mathbf{d}_{i}^1, \mathbf{d}_{i}^2, \mathbf{t}_{i}]$ is a frame predefined at the initial time. Its values are updated with a time matching scheme from time $t$ to time $t+\Delta t$, where $\Delta t$ is the time step size. Therefore, its values are known and we use it to measure the angle $\theta_i$ with the material frame $[\mathbf{m}_{i}^1, \mathbf{m}_{i}^2, \mathbf{t}_{i}]$.  Note that the third director in both the frames is the tangent along the edge, i.e., $\mathbf{t}_{i} = \frac{\mathbf{e}_{i}}{ \| \mathbf{e}_{i} \|}$, and the frames are orthonormal. The $4n+3$ sized degrees of freedom (DOF) vector representing the configuration of the elastic rod is $\mathbf{q} = [\mathbf{q}_0, \theta_0, \mathbf{q}_1, \ldots ,\theta_{n-1}, \mathbf{q}_n]$. As the robotic manipulator imposes boundary conditions on one end of the rod while the other end is clamped, the rod deforms. DER computes the DOF vector $\mathbf{q} (t)$ as a function of time by integrating the equation of motion (EOM) at each time stamp.

Before describing the EOM, we outline the elastic energies of a rod as a function of $\mathbf{q}$. The stretching energy is
\begin{equation}
    E_s = \sum_{i=0}^n \frac{1}{2} k_s \left( 1 - \frac{ \lVert \mathbf{e}_i \rVert}{\lVert \overline{\mathbf{e}_i} \rVert} \right)^2 \lVert \overline{\mathbf{e}_i} \rVert,
    \label{eq:stretching}
\end{equation}
where $k_s = EA$ is the axial stiffness, $E$ is Young's modulus, $A$ is the cross sectional area, and $\| \overline{\mathbf{e}_i} \|$ is the length of the undeformed edge $\overline{\mathbf{e}_i}$. In our case, the undeformed configuration is the initial configuration $\mathbf q(0)$. The bending energy is
\begin{equation}
    E_b = \sum_{i=1}^{n-1} \frac{1}{2} \frac{k_b}{\lVert \overline{\mathbf{e}_i} \rVert} \left( 2 \tan \frac{\phi_i}{2} - 2 \tan \frac{\phi^0_i}{2} \right)^2,
    \label{eq:bending}
\end{equation}
where $k_b = \frac{E \pi h^4}{4}$ is  the bending stiffness, $h$ is the rod radius, $\phi_i$ is the turning angle at a node (as shown in Fig.~\ref{fig::DER}(b)), and $\phi_i^0$ is the turning angle in the undeformed state. For the naturally straight rod studied here, $\phi_i^0 = 0$. The twisting energy is
\begin{equation}
    E_t = \frac{1}{2} \sum_{i=1}^{n-2} \frac{1}{\Vert e^i \Vert} k_t \tau_i^2,
    \label{eq::twisting}
\end{equation}
where $k_t = \frac{E \pi h^4}{4(1+v)}$ is the twisting stiffness, $\tau_i = \theta_i - \theta_{i-1} + \Delta \tau_i^{ref}$ is the discrete twist at node $\mathbf q_i$, and $\Delta \tau_i^{ref}$ is the reference twist, which is related to the twist between the reference frame on edge $\mathbf e_{i-1}$ and $\mathbf e_i$.

To march from time $t_i$ to $t_{i+1} = t_i + \Delta t$, the DER algorithm uses the implicit Euler method to integrate the EOM, which can be obtained via Newton's second law:
\begin{subequations}
	\begin{align}
	\begin{split}
	& \frac{\mathbb{M}}{\Delta t} \left( \frac{\mathbf{q} (t_{i+1}) - \mathbf{q} (t_i)}{\Delta t} - \dot{\mathbf{q}} (t_i) \right) = \mathbf{F}^{\textrm{int}} + \mathbf{F}^{\textrm{ext}},
	\label{eq::EOM1}
	\end{split}\\
	\begin{split}
	& \mathbf{F}^\textrm{int} = \frac{\partial (E_{s}+E_{b}+E_{t})}{\partial \mathbf{q}},
	\end{split}\\
	\begin{split}
	& \dot{\mathbf{q}} (t_{i+1}) = \frac{\mathbf{q} (t_{i+1}) - \mathbf{q} (t_i)}{\Delta t},
	\label{eq::EOM2}
	\end{split}
	\end{align}
	\label{eq4}
\end{subequations}
where $\mathbb{M}$ is the lumped mass matrix (which is a diagonal matrix of size $4n+3$), $F^\textrm{int}$ is the elastic force vector of size $4n+3$, and $F^\textrm{ext}$ is the external force vector of same size. Dot represents time derivative, i.e., $\dot{\mathbf{q}} (t_i)$ is the velocity vector of size $4n+3$ at time $t_i$. In the above time marching scheme, the {\em old} DOF vector $\mathbf{q} (t_i)$ and the old velocity vector $\dot{\mathbf{q}} (t_i)$ are known. Equation~\ref{eq::EOM1} is solved to obtain the {\em new} DOF vector $\mathbf{q} (t_{i+1})$. The new velocity vector is then simply calculated using equation~\ref{eq::EOM2}.


\subsection{Manipulation scheme to explore stable configurations}

We now propose a scheme for manipulating a helical rod that will incite instabilities, thereby allowing us to explore the points in the $\kappa$-$\tau$-$\omega$ parameter space corresponding to stable configurations.  As described earlier, it has been shown that each ray extending from the origin within the $\kappa$-$\tau$-$\omega$ parameter space intersects the boundary separating stable and unstable helices exactly once.  Moving along such a ray within the parameter space corresponds to continuously changing the helical rod's shape.  Therefore, we propose a manipulation scheme that begins with a straight untwisted rod (corresponding to the origin in the $\kappa$-$\tau$-$\omega$ parameter space), and then moves along a search direction $\mathbf{S}$ within this space.  We use $\lVert \mathbf{S} \rVert$ to denote the distance from the origin along the searching direction and $\overline{\mathbf{S}}$  to denote a unit vector in the search direction.  Beginning at the origin, we move along $\mathbf{S}$ in the $\kappa$-$\tau$-$\omega$ space until an instability is encountered.

\begin{figure}[t]
\centerline{\includegraphics[width =\columnwidth]{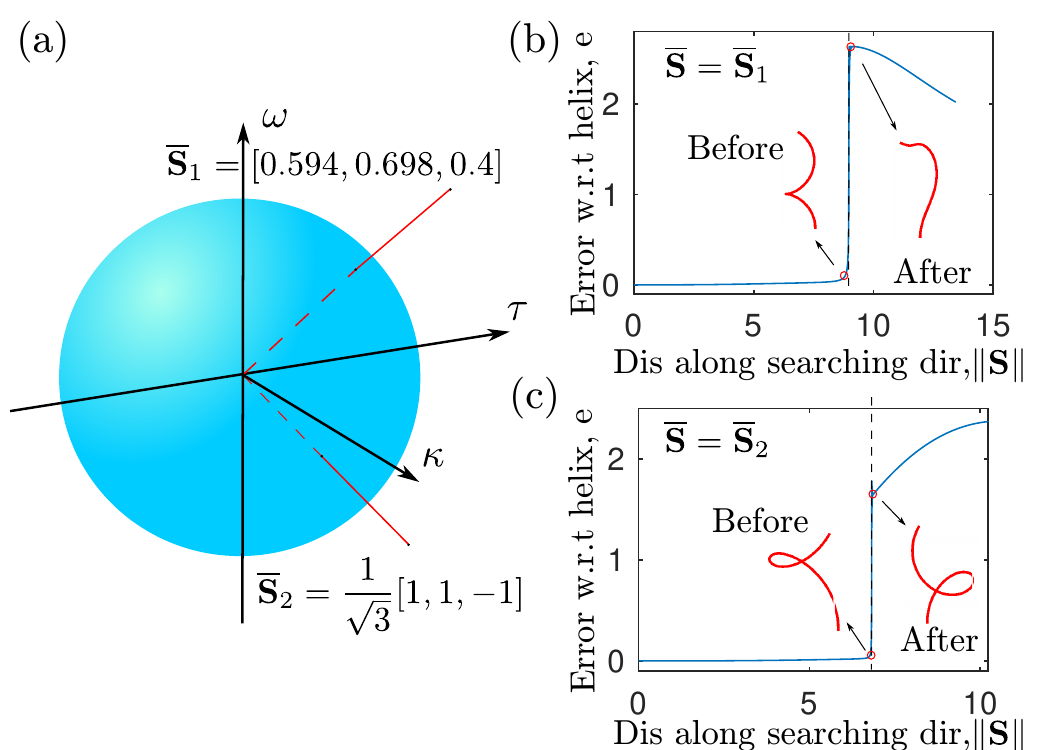}}
\caption{Scheme to incite an instability in a helical rod. (a) Parameterized space of helical configurations and two searching directions; (b) Error between the simulated rod and the expected helical shapes vs. distance along the searching direction $\lVert \mathbf{S} \rVert$ when $\overline{\mathbf{S}} = [0.594, 0.698, 0.4]$; (c) Error between the simulated rod and the expected helical shapes vs. distance along the searching direction $\lVert \mathbf{S} \rVert$ when $\overline{\mathbf{S}} = \frac{1}{\sqrt 3}[1, 1, -1]$;
}
\label{fig::scheme}
\end{figure}

Two examples of this procedure are shown in Fig.~\ref{fig::scheme}.  Fig.~\ref{fig::scheme}(a) shows the $\kappa$-$\tau$-$\omega$ parameter space along with a half-sphere of radius 1.  Each point on this half-sphere corresponds to a different search direction $\overline{\mathbf{S}}$, two of which are shown in the figure.  Moving along either search direction corresponds to manipulating a helical rod and continuously changing its curvature, torsion, and twisting moment.  This manipulation process was completed using the numerical simulation framework described in the previous section, and the results for the search directions $\overline{\mathbf{S}}_1$ and $\overline{\mathbf{S}}_2$ are shown in Fig.~\ref{fig::scheme}(b)-(c).  To measure the error between the simulated rod and the predicted helical shape, we calculate the average deviation of the rod configuration from the predicted helix, normalized by the helix's radius.  The expression for this error $e$ is given by
\begin{equation}
    e = \frac{1}{n+1}\sum_{i = 0}^{n} \frac{\kappa^2 + \tau^2 }{\kappa} \lVert \mathbf{q}_{i} - \mathbf{q}_{i}^\textrm{helix} \rVert,
    \label{eq::twisting}
\end{equation}
where $n+1$ is the number of discrete nodes representing the centerline of the rod, $\mathbf{q}_i$ is the position of the node, and $\mathbf{q}_i^\textrm{helix}$ is the position of the node when the rod is assumed to be helical.  In Fig.~\ref{fig::scheme}(b)-(c), we see that this error initially remains small and then jumps sharply at a critical distance along the search direction.  This sudden increase corresponds to an instability at which the rod jumps to another non-helical configuration.  This process was repeated for 58,352 search directions using numerical simulation, and the resulting points $\mathbf{S}$ at which instabilities occurred were recorded.  These points were then used to generate the surface in Fig.~\ref{fig::exp}, which represents the predicted boundary between stable and unstable configurations within the $\kappa$-$\tau$-$\omega$ parameter space.

\section{Robotic system}
\label{sec::robot}
In this section, we describe a robotic system that implements the manipulation scheme from the previous section within the $\kappa$-$\tau$-$\omega$ parameter space.  This system allows for automated stability testing and unsupervised collection of relevant data.  We first provide an overview of the robotic system.  We then discuss how boundary conditions for the rod corresponding to points in the $\kappa$-$\tau$-$\omega$ parameter space are implemented.  Finally, we describe the vision system used to detect instabilities, and we discuss the effects of disturbances in the robotic system.


\subsection{Overview of the robotic system}
\begin{figure}[t]
\centerline{\includegraphics[width =\columnwidth]{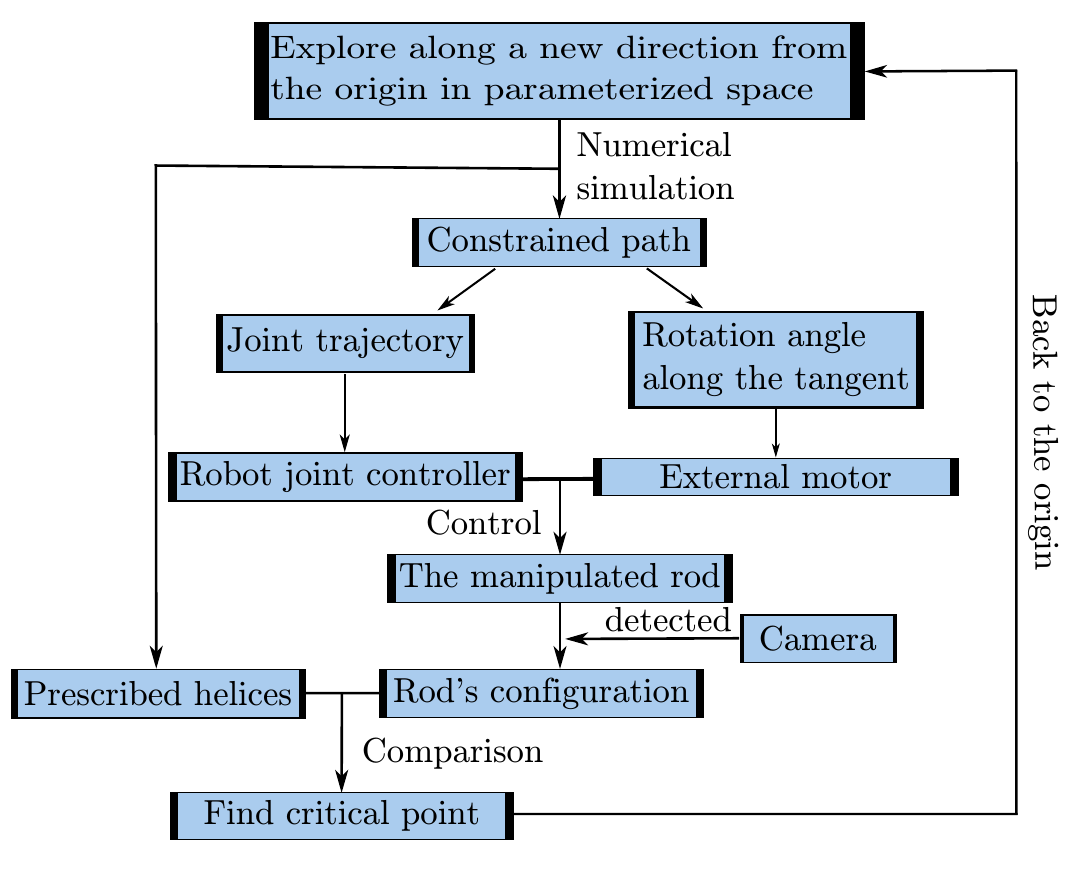}}
\caption{Flow chart of the manipulation scheme.}
\label{fig::flowChart}
\end{figure}

A flowchart of the robotic manipulation scheme is shown in Fig.~\ref{fig::flowChart}. The robotic system is composed of three parts: a collaborative robot, an externally mounted motor, and a camera. The collaborative robot imposes the prescribed position and tangent on one end of the manipulated rod while the mounted motor imposes the required torsion by rotating one end of the rod. In other words, the collaborative robot and the mounted motor work together to apply the required clamped boundary conditions on the manipulated rod. A camera is used to image the configuration of the rod undergoing manipulation. The images are used to calculate the difference between the experimental rod and the predicted helical shape when searching for the boundary of the set of stable helices. As described in Fig.~\ref{fig::scheme}, a large increase in the error between the manipulated rod and the predicted configuration indicates that a point on the boundary between stable and unstable configurations has been found.  The robotic system is able to explore the boundary of the set of stable helices by repeating this experimental procedure along different directions $\mathbf{S}$.

\subsection{Boundary conditions and path planning}
\begin{figure}[t]
\centering
\includegraphics[width =.8\columnwidth]{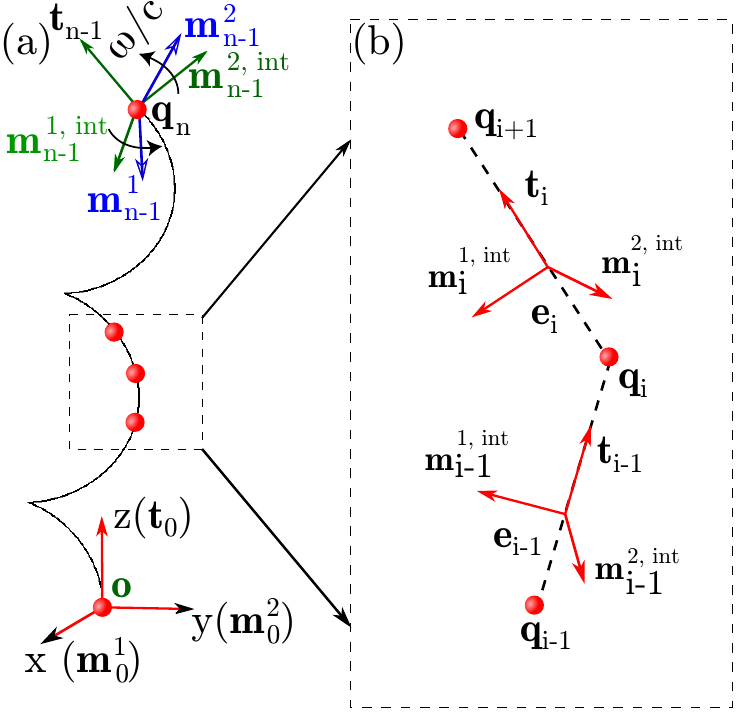}
\caption{(a) Illustration of the references frames on the two ends; (b) Schematic diagram of parallel transport.
}
\label{fig::BC}
\end{figure}

In this study, the boundary conditions on the two ends of the rod are clamped, which require fixing both the position and the tangent. The position can be expressed by three variables $[x, y, z]$ in the world coordinate system -- a reference frame fixed to the environment -- and the tangent -- a unit vector -- can be specified using a rotation matrix. First, we discuss our implementation of the conditions on the position using the robot. As shown in Fig.~\ref{fig::BC} (a), we regard one clamped end (the mounting end) of the rod as the origin of the world frame without any loss in generality. This position $\mathbf q_0$ is the first node in the discrete representation. For a helical shape, the last node $\mathbf q_n$ held by the end-effector of the robot can be expressed using the geometrical properties ($\kappa$ and $\tau$) of the helix such that
\begin{equation}
   \mathbf{q}_{n} =  \begin{bmatrix}
   - \frac{ \kappa \tau \left( -1 + (\kappa^2 + \tau^2)^{3/2} \sin{ \frac{1}{(\kappa^2 + \tau^2)^{3/2}}}\right)}{\kappa^2 + \tau^2} \\
   \kappa \left( \cos \left(\frac{1}{ (\kappa^2 + \tau^2)^{3/2}} \right) -1 \right) (\kappa^2+\tau^2)\\
   \frac{\tau^2}{\kappa^2+\tau^2} + \kappa^2 \sqrt{\left(\kappa^2+\tau^2 \right)} \sin \left( \frac{1}{(\kappa^2+\tau^2)^{3/2}}\right)
   \end{bmatrix}.
  \label{eq:position}
\end{equation}

Next, a method to implement the tangent or orientation boundary condition has to be developed.
The material frame $[\mathbf{m}_{n-1}^1, \mathbf{m}_{n-1}^2, \mathbf{t}_{n-1}]$ on the last edge (manipulated end) gives the required rotation matrix. Referring to Fig.~\ref{fig::flowChart}, this material frame is imposed by the joint controller of the end-effector and the external motor mounted on the end-effector. When the twisting moment is zero ($\omega=0$), we denote the material frame of the last edge to be $[\mathbf{m}_{n-1}^{1, \textrm{int}}, \mathbf{m}_{n-1}^{2, \textrm{int}}, \mathbf{t}_{n-1}]$. This ``intermediate" material frame is imposed by the joint controller of the robot. %
The intermediate material frame can be computed using parallel transport, which allows us to move the material frame from one edge to another without twisting about the tangent. Referring to Fig.~\ref{fig::BC} (b), given the material frame on edge $\mathbf e_{i-1}$, the material frame on $\mathbf e_i$ that does not generate any twist can be computed from the following steps.
\begin{equation}
\begin{aligned}
    \mathbf{b} &= \frac{\mathbf t_{i-1} \times \mathbf t_i }{ \lVert \mathbf t_{i-1} \times \mathbf t_i \rVert} \; \textrm{where} \; \mathbf t_{i-1} = \frac{\mathbf e_{i-1}}{\lVert \mathbf e_{i-1} \rVert}, \mathbf t_{i} = \frac{\mathbf e_{i}}{\lVert \mathbf e_{i} \rVert}, \\
    \mathbf{m}_i^{1, \textrm{int}} &= (\mathbf{m}_{i-1}^{1, \textrm{int}} \cdot (\mathbf{t}_{i-1} \times \mathbf{b})) (\mathbf{t}_i \times \mathbf{b}) + (\mathbf{m}_{i-1}^{1, \textrm{int}} \cdot \mathbf{b}) \mathbf{b},\\
    \mathbf{m}_i^{1, \textrm{int}} &= \frac{\mathbf{m}_i^{1, \textrm{int}}} {\lVert \mathbf{m}_i^{1, \textrm{int}} \rVert},\\
    \mathbf{m}_i^{2, \textrm{int}} &= \mathbf t_i \times \mathbf{m}_i^{1, \textrm{int}},
\end{aligned}
\label{eq:parrallelTrans}
\end{equation}
where $[\mathbf{m}_{i}^{1, \textrm{int}}, \mathbf{m}_{i}^{2, \textrm{int}}, \mathbf{t}_{i}]$ is the intermediate frame on the i-th edge with zero twist compared with the material frame $[\mathbf{m}_0^1, \mathbf{m}_0^2, \mathbf{t}_0]$ on the fixed end.
By sequentially parallel transporting the material frame from the first edge to the last one, we can obtain the intermediate frame $[\mathbf{m}_{n-1}^{1, \textrm{int}}, \mathbf{m}_{n-1}^{2, \textrm{int}}, \mathbf{t}_{n-1}]$.

This intermediate frame and the prescribed material frame $[\mathbf{m}_{n-1}^1, \mathbf{m}_{n-1}^2, \mathbf{t}_{n-1}]$ share the tangent $\mathbf{t}_{n-1}$ as the third director. Therefore, only a scalar quantity -- the rotation angle -- is needed to obtain the prescribed material frame from the intermediate frame. As indicated in Fig.~\ref{fig::flowChart}, an external motor that is mounted on the end-effector rotates the last edge by a rotation angle $\omega/c$, where $c$ is the ratio between twisting stiffness and bending stiffness. Fig.~\ref{fig::BC}(a) schematically shows the two frames and the rotation angle. The reason behind using an external motor to impose the rotation is that the rotation angle in this study can be so large that it falls outside the joint limits of the collaborative robot.

\begin{figure*}[t]
\centering
\includegraphics[width = 2\columnwidth]{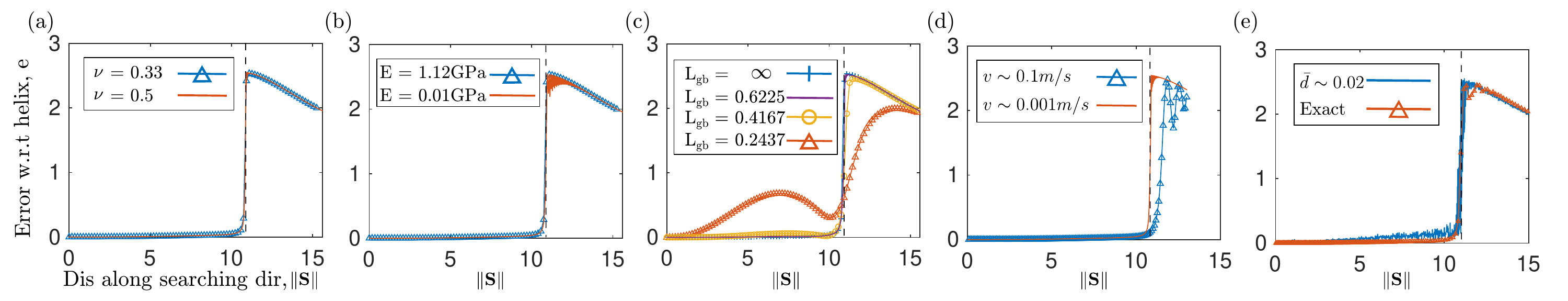}
\caption{The effects of (a) Poisson's ratio, (b) Young's Modulus, (c) the gravito-bending length, (d) manipulator velocity, and (e) jittering on the point of instability.  In each figure, the horizontal axis is the distance along the search direction $\overline{\mathbf{S}} = 1/\sqrt{3}[ 1 \ 1 \ 1]$, and the vertical axis is the error with respect to the expected helical shape.  The predicted point of instability is denoted by the vertical dashed line.
}
\label{fig::Dist}
\end{figure*}

Knowing the required boundary conditions of a series of helical shapes to be explored, we can construct a constrained path in Cartesian space. During motion planning, singular configurations of the robot should be avoided since these singularities could result in a high speed of the manipulator, thereby influencing the stability of the rod.  To minimize the likelihood of encountering these singularities, the desired path of the manipulator was divided into a series of discrete points and the Descartes planner~\cite{edwards2015descartes} was used to plan a corresponding path in the robot's joint space.  The collaborative robot used in our experiments has seven DOF, and there are multiple joint solutions for a specified pose along the desired path. Between discrete points along the path, the Descartes planner minimizes the function $f = \lVert \theta_{i+1}^r - \theta_{i}^r\rVert$, where $\theta^r$ is the robot joint solution and subscripts denote the index of the corresponding discrete point on the  path. The minimization of $f$ reduces the likelihood of encountering large and sudden changes in the robot's joint angles due to singularities.  During motion planning, we also account for self-collisions and joint limits of the robotic system, resulting in a joint path that does not have sudden jumps associated with singularities.  We note, however, that for a robotic system with fewer DOFs (e.g., 6 DOFs), avoiding singularities might be more challenging during motion planning.  In this circumstance, methods such as those described in~\cite{buss2004introduction} can be used during motion planning.

\subsection{Perception system}
The perception system is completed with a camera (Intel Realsense D435) that images the rod. Lightweight markers made of paper are attached along the rod (see Fig.~\ref{fig::fig1}) to track its configuration. Using the extrinsic and intrinsic matrices of the camera, the expected helical shapes from numerical simulations can be projected into the image domain of the experiments. The intrinsic camera matrix is provided by the vendor and the extrinsic matrix is measured with robot hand-eye calibration. The difference between the expected helical shape and the detected manipulated rod in the image domain is used to evaluate if an elastic instability occurs. When the elastic rod with a helical centerline reaches the critical point, it will snap into a non-helical shape and induce a large difference between the experimental shape and the prescribed helical shape. A representative example on this experiment vs. simulation comparison is shown in the supplementary video. The corresponding curvature $\kappa$, torsion $\omega$, and twist $\tau$ at the onset of elastic instability is a point on the boundary of the set of stable helical configurations. In this work, a detailed comparison between the 3D simulated configuration and the experimental configuration is not necessary; we are only interested in capturing the onset of instability.

\subsection{Effects of disturbances}

We now discuss potential sources of disturbances in the robotic system and their effects on the stability measurements.  First, in section~\ref{sec::ps}, we described how the parameters $\kappa$, $\tau$, $\omega$, and $c$ can be used to describe the set of all helical rod configurations, where curvature $\kappa$ and torsion $\tau$ are geometrical parameters of the rod that are independent of the rod's material. Furthermore, the ratio of twisting stiffness to bending stiffness, $c = k_t/k_b = 1/(1 + \nu)$, is only dependent upon the Poisson's ratio of the material. Varying $c$ results in a change of the rod's twisting strain, $\omega/c$, and a change of the rod's twisting energy. However, these two changes together result in no change in the rod's stability~\cite{borum2020helix}. To validate this result, we simulated a helical rod using the DER formulation along the search direction $\overline{\mathbf{S}} = 1/\sqrt{3}[ 1 \ 1 \ 1]$. The default parameters for these simulations were: Young's modulus $E = 1.12$ Gpa, Poisson's ratio $\nu = 0.33$, density $\rho = 1180$  kg/m$^3$, length $L = 1$ m, and radius $h = 0.781$ mm. In Fig.~\ref{fig::Dist} (a)-(b), we show the effect of varying Poisson's ratio and Young's modulus on the instability.  We see that the rod's material properties have a minimal effect on the instability point (the value of which is indicated by the vertical dashed line for the default parameters). Therefore, the material of the rod should have minimal influence on our stability results, and it is sufficient to use a single rod in the experiments.  In future work, our robotic system can be used to automate the experimental validation of the theoretical result regarding $c$ obtained in~\cite{borum2020helix}.

In addition to the rod's material, the effect of gravity on the rod must also be considered.  In this paper, we assume that the rod is sufficiently stiff so that gravity can be neglected.  To quantify this assumption, we use the gravito-bending length $L_{gb} = \left(\frac{h^2 E}{8 \rho g}\right)^{1/3}$, where $g$ is the acceleration of gravity~\cite{jawed2014coiling}.  $L_{gb}$ describes the balance between gravitational and bending energy, and the effects of gravity diminish as $L_{gb}$ increases.  Fig.~\ref{fig::Dist} (c) shows the effect of varying $L_{gb}$ on the instability, and we see that gravity becomes negligible when $L_{gb}/L > 0.6$.

The main external disturbances (i.e., external to the rod) can be divided into two components: the speed of the robotic manipulator and jittering of the manipulator. To assess these effects, we again conducted simulations with the default parameters described above.  Fig.~\ref{fig::Dist} (d) shows how varying the manipulator's speed $v$ affects the instability point by introducing inertial effects.  We conclude that the manipulator's speed should be sufficiently small so as to minimize the influence of these inertial disturbances. We also used the simulator to explore the effects of jittering, i.e., small deviations from the desired path. As shown in Fig.~\ref{fig::Dist} (e), we found that jittering of magnitude less than $d \backsim 0.02L$, where $L$ is the rod's length, has minimal influence on the rod's stability.

\section{Experiments and analysis}
\label{sec::exp}

The robotic system, manipulation scheme, and vision system described in section \ref{sec::robot} were implemented to conduct automated experimental testing of stability for helical elastic rods.  We used a collaborative robot (Sawyer, Rethink robotics) with seven degrees of freedom. The motor used for applying external torque is a stepper motor (NEMA 17) controlled with a microcontroller (Arduino Uno Rev3), and the camera used in the perception system is an RGBD camera (Intel Realsense D435). The elastic rod used was a superelastic nitinol wire with length $L = 0.5$ m, diameter $2r = 1.5875$ mm, Poisson's ratio $\nu = 0.33$, density $\rho = 6450$ $kg/m^3$, and Young's modulus $E = 67.5$ GPa.  For this rod, we have $L_{gb} / L = 1.3816$, and we can therefore neglect the effects of gravity.  Furthermore, the reported accuracy of the Sawyer robot is less than 0.01 cm, which is within the jittering tolerance established in the previous section.

As described in Ref.~\cite{borum2020helix}, there is a symmetry in the $\kappa$-$\tau$-$\omega$ parameter space between points located at $(\kappa,\tau,\omega)$ and $(\kappa,-\tau,-\omega)$.  We therefore only consider search directions with $\tau>0$, and we reflect these points on the stability boundary to generate data for $\tau<0$.  In future work, experiments will be conducted for all values of $\tau$ to validate this symmetry.  The rod was manipulated along a total of 328 search direction until an instability occurred, and the resulting values of $\kappa$, $\tau$, and $\omega$ at the instability were recorded.

\begin{figure}[t!]
\centerline{\includegraphics[width =\columnwidth]{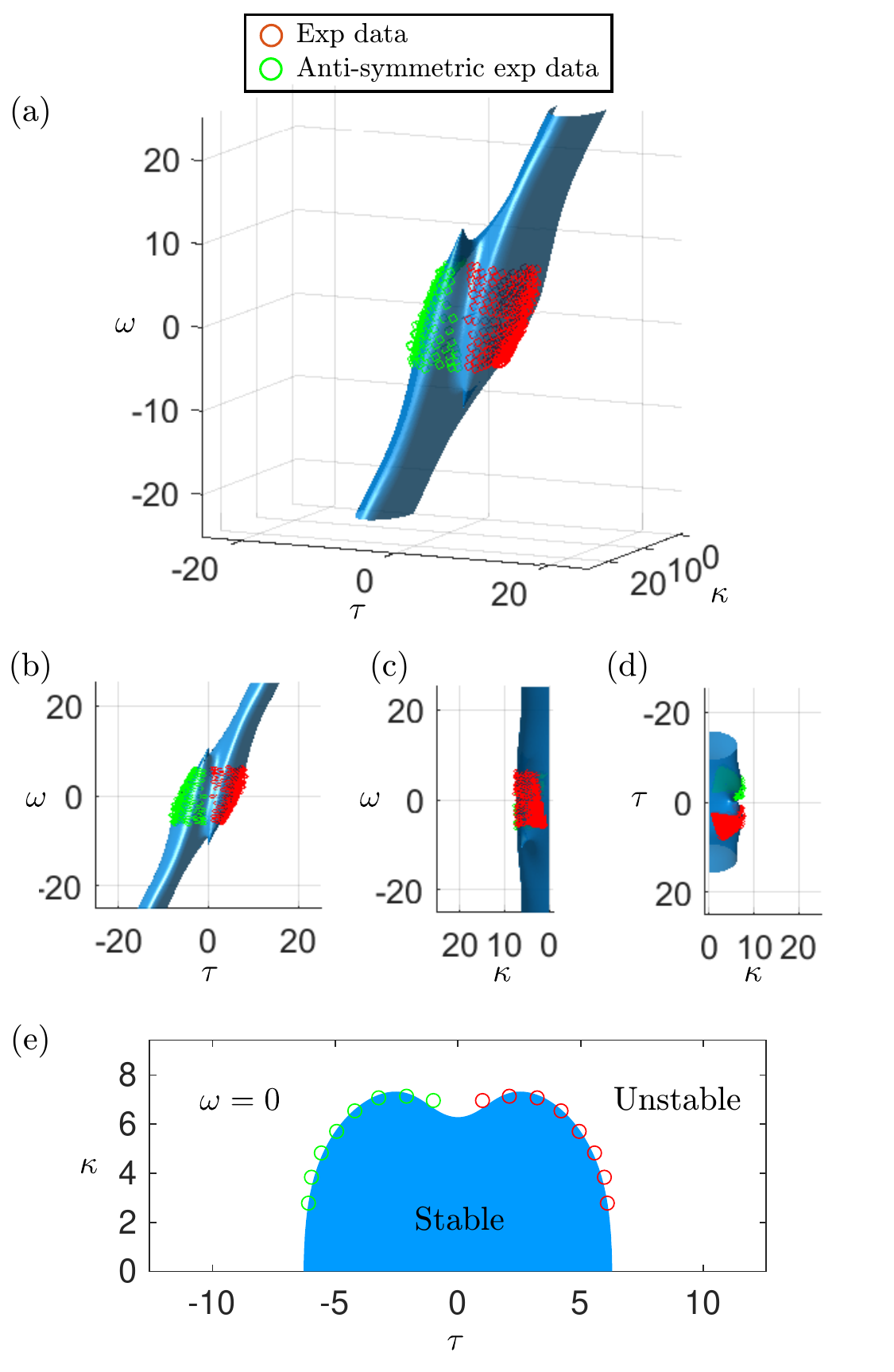}}
\caption{Comparison between simulation data (blue surface) and experiment data: red circles denote the experimental data; green circles denote the anti-symmetric experimental data. (a) Comparison between simulation results and experimental results; (b) View via negative $\kappa$ axis of (a); (c) View via negative $\tau$ axis of (a); (d) View via negative $\omega$ axis of (a); (e) The comparison between simulation and experimental results when $\omega = 0$.
}
\label{fig::exp}
\end{figure}

The blue surface in Fig.~\ref{fig::exp} (a) shows the predicted boundary between stable and unstable helical configurations within the $\kappa$-$\tau$-$\omega$ parameter space based on the DER numerical simulations.  The experimental data collected by our automated testing procedure is shown in red, and the anti-symmetric data generated from the symmetry is shown in green.  Views along each coordinate axis are shown in Fig.~\ref{fig::exp} (b)-(d), and Fig.~\ref{fig::exp} (e) shows the comparison for the two-dimensional section with $\omega = 0$, i.e., twist-free helices.  Our results show good agreement between the simulated and experimental data, suggesting that our automated testing procedure is able to accurately reproduce the stability boundary.

To quantify the difference between the simulated and experimental data, we introduce an error that measures the relative distance between points on the simulated and experimental stability boundaries along the same search direction.  Along a given search direction $\overline{\mathbf{S}}$, we let $\mathbf{S_s}=[\kappa_s, \tau_s, \omega_s]$ denote the resulting point on the stability boundary based on numerical simulations, and similarly we let $\mathbf{S_e}=[\kappa_e, \tau_e, \omega_e]$ denote the resulting point on the stability boundary based on the experimental data.  The relative error along the search direction $\overline{\mathbf{S}}$ is then defined as $err = \left\lvert \left( \lVert S_e \rVert - \lVert S_s \rVert \right) / \lVert S_s \rVert \right\rvert$.  The average error of our experimental results over all 328 search directions was 0.0272.  The maximum error found was 0.1298, and the standard deviation of the error was 0.0238. Based on these results, we can conclude that our automated experimental testing method can accurately determine when an elastic rod in a helical configuration loses stability.


\section{Conclusion}
\label{sec::conclusion}
This paper developed an automated testing procedure for determining the stability of a helical elastic rod.  Due to the repetitive nature of the test and the need to simultaneously manipulate both the position and orientation at one end of the rod, the robotic system was a key component of our testing procedure.  Experimental observations were compared with results from numerical simulations based on the DER algorithm, and their agreement suggests our method is able to accurately capture the onset of instabilities in helical rods.  Although our manipulation scheme relied on specific properties of helical rods, the other components of our testing method can be applied to analyze stability of non-helical rod configurations.


While our automated testing system was able to accurately reproduce the boundary between stable and unstable helical rods, there are areas in which the method could be improved.  First, a primary error source in our system was the stepper motor, due to its inability to apply a consistently linear rotation and latency when communicating with the robotic system.  Furthermore, since the position of one end of the rod as fixed in our experiment, self-collisions of the robot prevented us from exploring certain regions of the $\kappa$-$\tau$-$\omega$ parameter space. Using two robots to collaboratively manipulate the rod would provide additional dexterity, allowing us to explore a larger region of the parameter space. Our future work will focus on understanding how the robot's work space constrains the rod's mechanical parameter space.  Despite these areas for improvement, the methods described in this paper provide a foundation for using robotic systems to perform research in experimental mechanics.  In future work, a robotic system could be used to not only perform mechanics experiments, but to apply machine learning based on the collected data to improve the mechanical model of the object being manipulated.

\printbibliography

\end{document}